\PassOptionsToPackage{numbers}{natbib}
\PassOptionsToPackage{sort}{natbib}
\documentclass{article}

    \usepackage[final]{neurips_2020}

\usepackage[utf8]{inputenc} 
\usepackage[T1]{fontenc}    
\usepackage{hyperref}       
\usepackage{url}            
\usepackage{booktabs}       
\usepackage{amsfonts}       
\usepackage{nicefrac}       
\usepackage{microtype}      
\usepackage{xcolor}         
\usepackage{amsmath}
\usepackage{mathtools}
\usepackage{commands}
\usepackage{multirow}
\usepackage{tablefootnote}
\usepackage{caption}

\title{Is Support Set Diversity Necessary for Meta-Learning?}

\author{%
Amrith Setlur$^{\S}$\thanks{Authors contributed equally to this paper.} \qquad Oscar Li$^{\dagger*}$ \qquad Virginia Smith$^{\dagger}$ \\
    ${^\S}$Language Technologies Institute \qquad $^\dagger$Machine Learning Department\\ 
    Carnegie Mellon University \\
    \texttt{asetlur@cs.cmu.edu}\quad \texttt{oscarli@cmu.edu}     \quad \texttt{smithv@cmu.edu}
}

\begin{document}

\maketitle

\begin{abstract}
Meta-learning is a popular framework for learning with limited data in which an algorithm is produced by training over multiple few-shot learning tasks. For classification problems, these tasks are typically constructed by sampling a small number of support and query examples from a subset of the classes. While conventional wisdom is that task diversity should improve the performance of meta-learning, in this work we find evidence to the contrary: we propose a modification to traditional meta-learning approaches in which we keep the support sets fixed across tasks, thus reducing task diversity. Surprisingly, we find that not only does this modification \textit{not} result in adverse effects, it almost always improves the performance for a variety of datasets and meta-learning methods. We also provide several initial analyses to understand this phenomenon. Our work\footnote{Implementation available at: \url{https://github.com/ars22/fixml}} serves to: (i) more closely investigate the effect of support set construction for the problem of meta-learning, and (ii) suggest a simple, general, and competitive baseline for few-shot learning. 
\end{abstract}

\section{Introduction}


The ability to learn from limited experience is a defining aspect of human intelligence. 
The domain of {few-shot learning} aims to evaluate this criterion within machine learning~\cite{fei2006one}. 
For few-shot learning problems, meta-learning~\cite{thrun1998learning} techniques have attracted increasing attention.
Meta-learning methods typically construct few-shot classification tasks in an {episodic} manner by sampling \textit{support}  ($S$) and \textit{query}  ($Q$) examples from a fixed number of classes. The objective is then to deliver an {algorithm} that can perform well on the query points of a task by training on only a few support samples.

Conventional wisdom is that the performance of meta-learning methods will improve as we train on more diverse tasks. 
Therefore, during meta-training, we typically allow a task to be constructed from any possible pair of support and query sets. In this work, we question this notion, specifically investigating the effect that support set diversity has on meta-learning. 
Surprisingly, we find that reducing the total number of unique support sets not only does \textit{not} result in adverse effects---in most cases, it in fact yields significant performance improvements. 

The main contributions of this work are as follows: (1) We propose a simple modification to the meta-learning objective that applies to a broad set of methods, in which we aim to optimize a biased  objective by restricting the number of support sets used to construct tasks. (2) On multiple datasets and model architectures we empirically demonstrate the performance improvement of the learned algorithm when trained with our proposed objective. (3) Finally, we explore a framework to understand this surprising phenomenon and attempt to provide some insights into the observed gains. Our work delivers a simple, competitive baseline for few-shot learning and, more generally, investigates the role that task construction plays in meta-learning. 








\section{Background \& Related Work}

Meta-learning formulations  typically rely on episodic training, wherein an {algorithm} is learned to {adapt} to a task, given its support set, so as to minimize the loss incurred on the query set. Meta-learning methods differ in terms of the algorithms they learn: \textit{Gradient-based meta-learning methods} \cite{finn2017meta,finn2018probabilistic,rajeswaran2019meta,sung2018learning, lee2019meta} consider learning gradient-based algorithms, which are typically parameterized by an initialization (possibly together with some pre-conditioning matrix).
Recently, \textit{last-layer meta-learning methods} ~\cite{lee2019meta,oreshkin2018tadam,bertinetto2018metalearning, snell2017prototypical} have gained popularity. These methods don't learn an optimal model parameter initialization for the task distribution but instead learn a feature extractor for the covariates in the support and query sets. The extracted support set features along with their labels are used to learn the parameters of the last layer, e.g., in a non-parametric fashion \cite{snell2017prototypical} or by finding the unique solution for convex problems ~\cite{lee2019meta,bertinetto2018meta}.
  
Several recent works \cite{chen2019closer, chen2020new, mangla2020charting} on few-shot learning have proposed methods that don't directly optimize the original meta-learning objective. Instead they chose to minimize a \textit{transfer-learning} based loss where a deep network is trained in a supervised manner on all the $(x, y)$ pairs present in the dataset. In contrast to episodic training, they show that a model trained without any notion of a task can be used to construct an algorithm whose performance matches or improves upon meta-learning methods across many benchmarks. 
Similar to these works, we propose to optimize an alternate objective for meta-learning; however, our work differs by retaining the episodic structure from the meta-learning framework but restricting the support set (and by consequence task) diversity. 

In addition to using other objectives for few-shot learning,  recent work aims to improve meta-learning by explicitly looking at the task structure and their relationships. Among these, \citet{yin2019meta}  propose to handle the lack of mutual exclusiveness among different tasks through an information-theoretic regularized objective. \citet{collins2020task}  propose a minimax objective to make the learned algorithm task-robust, which improves the performance of the algorithm on the most difficult task in the face of a task pool with diverse difficulty. \citet{liu2020task} propose to augment the set of possible tasks by augmenting the pre-defined set of classes that generate the tasks with 
varying degrees of rotated inputs as new classes. In addition, several popular meta-learning methods \cite{snell2017prototypical, lee2019meta}, in order to improve the meta-test performance, change the number of ways or shots of the sampled meta-training tasks, thus increasing the complexity and diversity of the tasks. In contrast to these works, we look at the structure and diversity of tasks specifically through the lens of support set diversity, and show that, surprisingly, reducing diversity (by fixing support set) not only maintains---but in many cases significantly improves---the performance of meta-learning. Our results indicate that this simple modification to meta-learning is an effective method worthy of future study. 

\section{Meta-Learning with a Fixed Support Pool}
\label{sec:framework}

We begin with some notation that we will use to re-formulate the meta-learning objective. This will allow us to introduce the concept of a \textit{support pool} in a manner amenable with current meta-learning methods.

\textbf{Notation.} 
In a standard offline meta supervised learning problem, we are given a dataset $\calD \subseteq \mathcal{X} \times [N]$ with examples from $[N] \coloneqq 1, \ldots, N$ different classes. A task $(C, S, Q)$ is constructed by first choosing a subset of classes $C \subseteq [N]$, and conditioning on $C$, a support set $S$ and a query set $Q$ are sampled i.i.d. from  $\calD^{C} \coloneqq \{(x, y) \in \calD: y \in C\}$. The goal is to learn  the parameters of an algorithm $w \in \Real^d$ so that when given a support set $S$, the algorithm will output a model that achieves a low loss value on the query set. We denote this loss value as a function of the algorithm parameter, support, and query set: $\ell: \Real^d \times \cbrck{S} \times \cbrck{Q} \to \Real.$ To identify the optimal parameter of the algorithm $w$, we optimize the objective in [A] below:
\begin{align}
\begin{split}
    \textrm{[A]:}\, \min_w \E_C \E_{(S, Q) \; \mid \; C} \ell(w, S, Q)    
\quad \displaystyle \equiv \quad
\textrm{[B]:}\, \min_w \E_{S_p \sim \Unif(\calP)} \E_{C} \E_{Q \;\mid\; C} \ell(w; S_p^C, Q)
\label{eq:ML}
    \end{split}
\end{align}

\textbf{Reformulating the meta-learning objective.} 
We now give another view of this objective, which will motivate our proposed modified objective. Instead of sampling from the set of classes $C$ when constructing a task, we consider first sampling uniformly from $\calP$,       a collection of support pools. A \textbf{\textit{support pool}} $S_p \in \calP$ is a smaller subset of the entire dataset $D$ that contains some examples from every class in $D$, \ie $\cbrck{y: (x, y) \in S_p}=[N]$. With $S_p$, we sample a subset of classes $C \subseteq [N]$ as before and take all the examples from $S_p$ that are of the classes in $C$. This set of examples is now our support set which we denote by $S_p^C = \cbrck{(x, y) \in S_p: y \in C}$. With the classes $C$ determined, we can then sample the query set $Q$ of examples from class $C$ the same way as described previously. In this view, the marginal distribution of the support query set pair $(S_p^C, Q)$ is exactly the same as that of $(S, Q)$ discussed before. As a result, we can rewrite our meta-learning objective as in \eqref{eq:ML}[B]. 

This reformulation prompts us to first check how big the collection of support pools $\calP$ typically is. In the most unconstrained form, $\calP$ can contain all possible subsets of a dataset where we have at least one sample from every class $(\forall c \in [N])$. However, most meta-learning methods only train on episodic tasks where the support $S$ is of a specific configuration, i.e., the support of each task is comprised of exactly $n$ \textit{ways} (classes) with each class having $k$ \textit{shots} (examples). 
For instance, when training on \textit{miniImagenet} \citep{vinyals2016matching} $5$-way $5$-shot ($n=5, k=5$) tasks, all tasks that are sampled consist of exactly $5$ shots from each of the $5$ classes. Hence, it also makes sense to limit $\calP$ to only support pools that obey a certain structure. In this work, we consider $\calP$ to consist only of subsets of $\calD$ that have exactly $k$ samples from each class i.e.
$\calP = \{S_p: {\footnotesize \sum}_{(x, y)\in S_p} \I(y = c)=k,  \;\forall c \in [N] \}$. 
The meta-training set of miniImagenet \citep{ravi2016optimization} has 64 classes with 600 examples each, 
resulting in {\scriptsize $|\calP| = \binom{600}{5}^{64} \approx 3\times 10^{755}$} support pools. 
This makes us wonder: \textit{Is this astronomically large number of support pools necessary or possibly redundant to achieve good meta-test performance? Can we use a much smaller number of support pools instead?}

\textbf{Fixing the support pool.} Motivated from the questions above, we take the most extreme reduction approach: before training begins, we randomly construct one support pool $S_{p,0}$ and let $\calP = \cbrck{S_{p, 0}}$ (just this single support pool) and solve for the corresponding optimization problem:
\begin{align}
\min_w \E_{C} \E_{Q \;\mid\; C} \ell(w; S_{p,0}^C, Q).
\label{eq:fixML}
\end{align}
Note that the expectation over $S_p$ is removed because there is only a single possible realization of the random support pool. In terms of the total number of different \textit{possible support sets} that the algorithm being learned can see, we have also achieved a reduction factor of {\scriptsize $\binom{600}{5}^{5}$} times for the aforementioned miniImagenet example. In the rest of the paper, we first show the performance obtained by optimizing this reduced objective \eqref{eq:fixML} for some of the best-performing meta-learning methods on multiple meta-learning problems and/or with different model architectures. We then make an initial attempt at understanding the effectiveness of using this fixed support pool objective.

\section{Experiments}
\label{sec:experiments}


We now aim to understand the differences in generalization performance between the algorithm learned by optimizing~\eqref{eq:fixML}, i.e. our proposed objective (which we denote \fixml), and the algorithm obtained by optimizing the original meta-learning objective \eqref{eq:ML} (which we denote \ml). 

For every \fixml experiment, we randomly sample $k$ samples (shots) from each class in the dataset to form the support pool $S_{p,0}$, which is then fixed throughout training. Here, different runs of \fixml on the exact same problem and meta-learning method can use different (but fixed) support pools. Later in this section, we  will discuss how this randomness affects meta-test performance over multiple \fixml runs. For an unbiased estimation and a fair comparison, when we evaluate a \fixml learned algorithm on the meta-validation or meta-test set, we compute the loss and accuracy using the objective given in \eqref{eq:ML} (i.e., an average over all possible support pools), just as we would for the \ml learned algorithm.

To fully evaluate the performance differences, we compare \fixml and \ml along three axes: 
\vspace{-.25em}
\begin{enumerate}[label=\textbf{\arabic*}.,wide, labelwidth=!, labelindent=0pt,itemsep=1mm]
\item \textbf{Meta-learning methods:} We focus our initial experiments on \textit{protoypical-networks} (PN) \cite{snell2017prototypical} which uses a non-parametric last layer solver, as \textit{Protonets} is a highly competitive \ml approach, and also has the benefit of training faster and with less memory than competitors such as \textit{SVM} \cite{lee2019meta}, \textit{Ridge Regression} (RR) \cite{bertinetto2018metalearning}.
However, we also experiment with SVM and RR-based solvers to confirm that the trends observed with Protonets generalize to these methods. As initialization-based methods are harder to optimize for highly over-parameterized models, we defer such evaluations to future work.   

\item \textbf{Datasets:} Our study mainly uses two of the most widely-used benchmarks in few-shot learning: (i) \textit{miniImagenet} (\mini) \citep{vinyals2016matching}, which consists of 100 classes of $84 \times 84$ images, split into 64 train, 16 validation and 20 test, and (ii) \textit{CIFAR-FS} (\cif) \cite{bertinetto2018metalearning} with an identical split but with $32 \times 32$ images. 
Additionally for the SVM method we 
elect to compare on the more difficult \textit{FC-100} (\FC) dataset, as the SVM-based solver has achieved significantly superior performance on this benchmark \cite{lee2019meta}.

\item \textbf{Architectures:} We conduct our main experiments using \textit{Resnet-12}, as many meta-learning methods \cite{bertinetto2018meta,oreshkin2018tadam} have relied on this architecture to achieve good performance. Additionally, we experiment with another wider but shallower over-parameterized architecture, \textit{WideResNet-16-10} \citep{zagoruyko2016wide} (also used by \cite{rusu2018meta}), and \textit{Conv64}, which was used by some initial works in meta-learning \citep{finn2017meta, snell2017prototypical}.
\end{enumerate}


\begin{figure}
    \centering
    \includegraphics[width=0.99\linewidth]{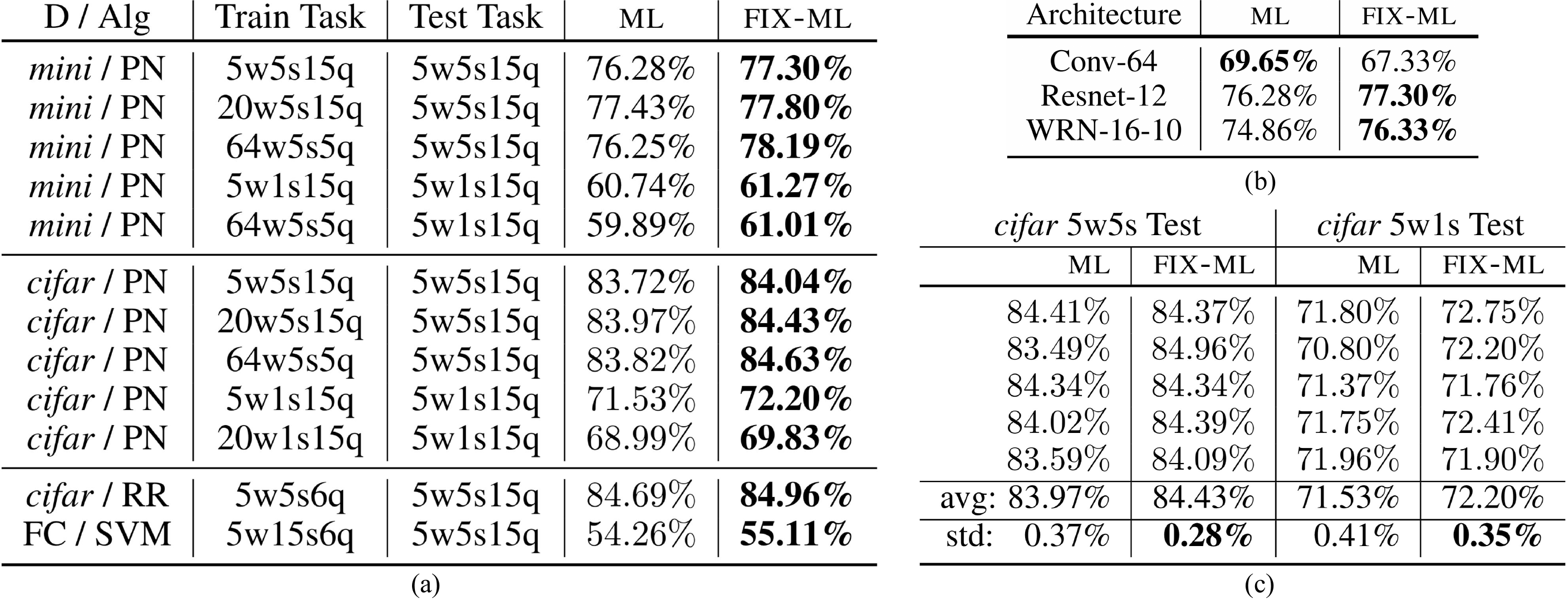}
    \caption{Performance gains observed upon reducing support set diversity by fixing a single support pool. In (a) we compare \fixml and \ml across datasets, algorithms and train/test configurations (XwYsZq $\coloneqq$ X-ways, Y-shots and Z-query), in (b) we evaluate the architecture's role in a comparison between \ml and \fixml on \textit{mini}-5w5s tasks and in (c) we show how the meta-test performance varies across runs for both \fixml and \ml.\vspace{1em}}
    \label{fig:fixml-comparison}
\end{figure}

\vspace{-0.5em}
\paragraph{Main Results.}
Figure \ref{fig:fixml-comparison}(a) consolidates the results of our experiments on different datasets with a myriad of training task configurations. Surprisingly, contrary to our intuition about task diversity, we can clearly see that optimizing the \textsc{fix-ml} objective performs similarly or in many cases better with respect to its \textsc{ml} counterpart.
In all the cases except for the \cif dataset, we evaluate the validation and test accuracies on $2000$ randomly drawn tasks consisting of novel classes and for each of them we observe a $95\%$ confidence interval of $\pm (0.32-0.36) \%$ around their respective means. For \cif, since the performance improvements are lower, we evaluate on $10000$ tasks instead and observe a $95\%$ interval of $\pm (0.13-0.14) \%$ (around the reported mean) which confirms the significance of the improvement.

Specifically, we see that on all of the \mini and \cif 5w1s and 5w5s evaluations, Protonets trained with \fixml achieve better performance. In addition, for each of these evaluations, we consider training with different number of ways and shots as proposed in \cite{snell2017prototypical, lee2019meta}. Our results also confirm that \fixml is better for a range of meta-training specifics. Besides, for other meta-learning methods like SVM and RR, fixing the support pool can also improve performance. For all the experiments in Figure~\ref{fig:fixml-comparison}(a) we use the Resnet-12 backbone. In order to further validate our findings, we show \textit{mini}-5w5s meta test results on two other architectures in Figure~\ref{fig:fixml-comparison}(b). WRN-16-10 has roughly 50\% more parameters than Resnet-12 and we see that \fixml continues to perform better than \ml in this over-parameterized regime. On the other hand, it seems that fixing the support pool for shallower backbones like Conv64 hinders the meta-test performance. For this we hypothesize that \fixml is effective for over-parameterized models and algorithms but less so for shallower models with fewer parameters. Further analysis on how model architecture and over-parameterization affects \fixml is a future research direction.

Recall that earlier in this section, we suspected that fixing the support pool randomly prior to the start of training may induce higher variance in the final meta-test performance over different runs.
From the observations in Figure \ref{fig:fixml-comparison}(c), we confirm that this is \textbf{not} true. We compare \fixml and \ml on \cif-5w5s and \cif-5w1s over five different runs each, where for every run of \fixml a different random support pool was sampled and then fixed. We can clearly see that the performance standard deviation is not only \textbf{not worse} but actually \textbf{better} than \ml. 
Finally, in Figure~\ref{fig:2}(a) we see that Protonets trained using \fixml objective achieves competitive results on \mini-5w5s when compared with other other meta-learning methods and with a simple transfer learning method \cite{chen2019closer} devoid of any self-supervised learning \cite{mangla2020charting} tricks or a transductive setting \cite{hu2020leveraging}.

 \paragraph{Implementation.} 
 We follow the learning rate schedule and data augmentation scheme of \cite{lee2019meta} 
 and use a task batch size of 4-8 when training with lower number of ways (e.g., $5$) due to observed improvements in convergence and generalization. 
 Additionally, we experiment with varying number of query points per class (in a single task). We observe that for some cases in Figure~\ref{fig:fixml-comparison}(a) this further improves the advantage of \fixml over \ml.

\vspace{-0.5em}

\section{Discussion}
\label{sec:discussion}

To understand the improved meta-test performance of the \fixml-learned algorithms,
 we {first} analyze their performance \textbf{on the meta-training set}. Evaluating the \fixml-learned algorithm's performance averaged over all possible support pools (equivalent to the standard \ml training loss) is important because, by the  generalization theory using uniform convergence bounds (e.g. VC dimension, Rademacher complexity \cite{bartlett2002rademacher}), the test loss tracks training loss more closely with more training samples.
In the case of meta-learning, the ``samples'' are tasks in the form of $(S,Q)$ pairs. Therefore, we see that an algorithm's \ml-training loss which considers all possible $(S,Q)$ pairs, 
should track the algorithm's loss on the meta-test set more reliably than the \fixml training loss which uses only a single support pool (inducing much fewer $(S,Q)$ pairs).

\begin{figure}[t] 
  \begin{minipage}[b]{0.4\linewidth}
  \centering
    {\scriptsize
    \begin{tabular}{ c | c | c  }
        Method &  Backbone & \textit{mini} 5w5s \\ \hline
        MatchingNet \cite{vinyals2016matching}  & Conv64 & 63.48\% \\
        RelationNet \cite{sung2018learning}  & Conv64 &  66.60\%  \\
        DCO  \cite{bertinetto2018meta} &  Conv64 &  68.40\% \\
        LEO \cite{rusu2018meta}  & WRN-28-10 & 77.59\%  \\
        \citet{chen2019closer} & Resnet-10 & 75.90\% \\
        TADAM   \cite{oreshkin2018tadam}     & Resnet-12  & 76.70\% \\
        MetaOptNet-SVM \cite{lee2019meta}  & Resnet-12 &  77.40\% \\ \midrule
        Protonet    & Resnet-12 &   77.43\%    \\
        Protonet \fixml & Resnet-12 &  \textbf{78.19\%}   \\
    \end{tabular}}
    \centering
    \caption*{(a)}
  \end{minipage}
  \begin{minipage}[b]{0.29\linewidth}
    \centering
    \includegraphics[width=\linewidth]{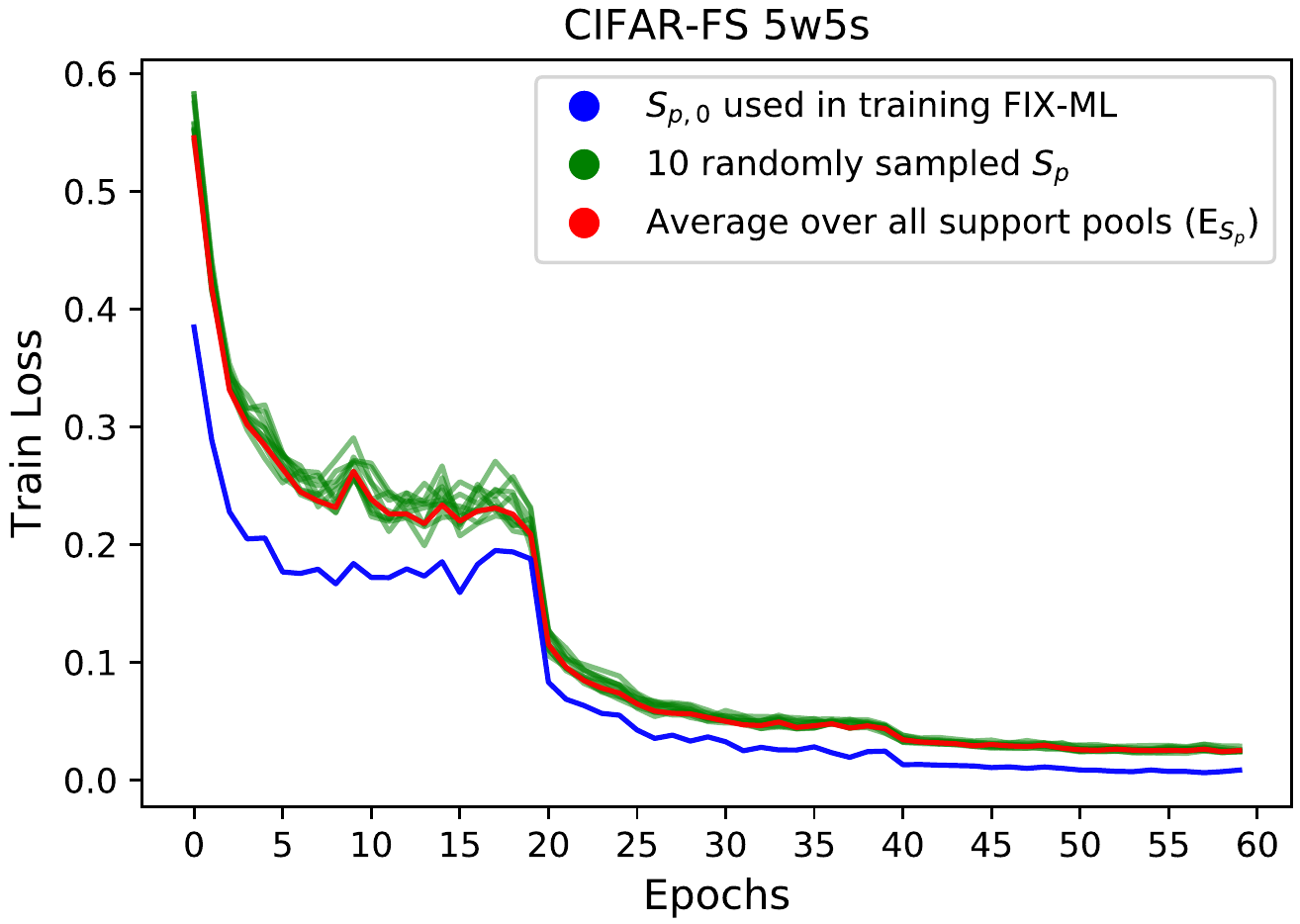} 
    \caption*{(b)}
  \end{minipage} 
  \begin{minipage}[b]{0.29\linewidth}
    \centering
    \includegraphics[width=\linewidth]{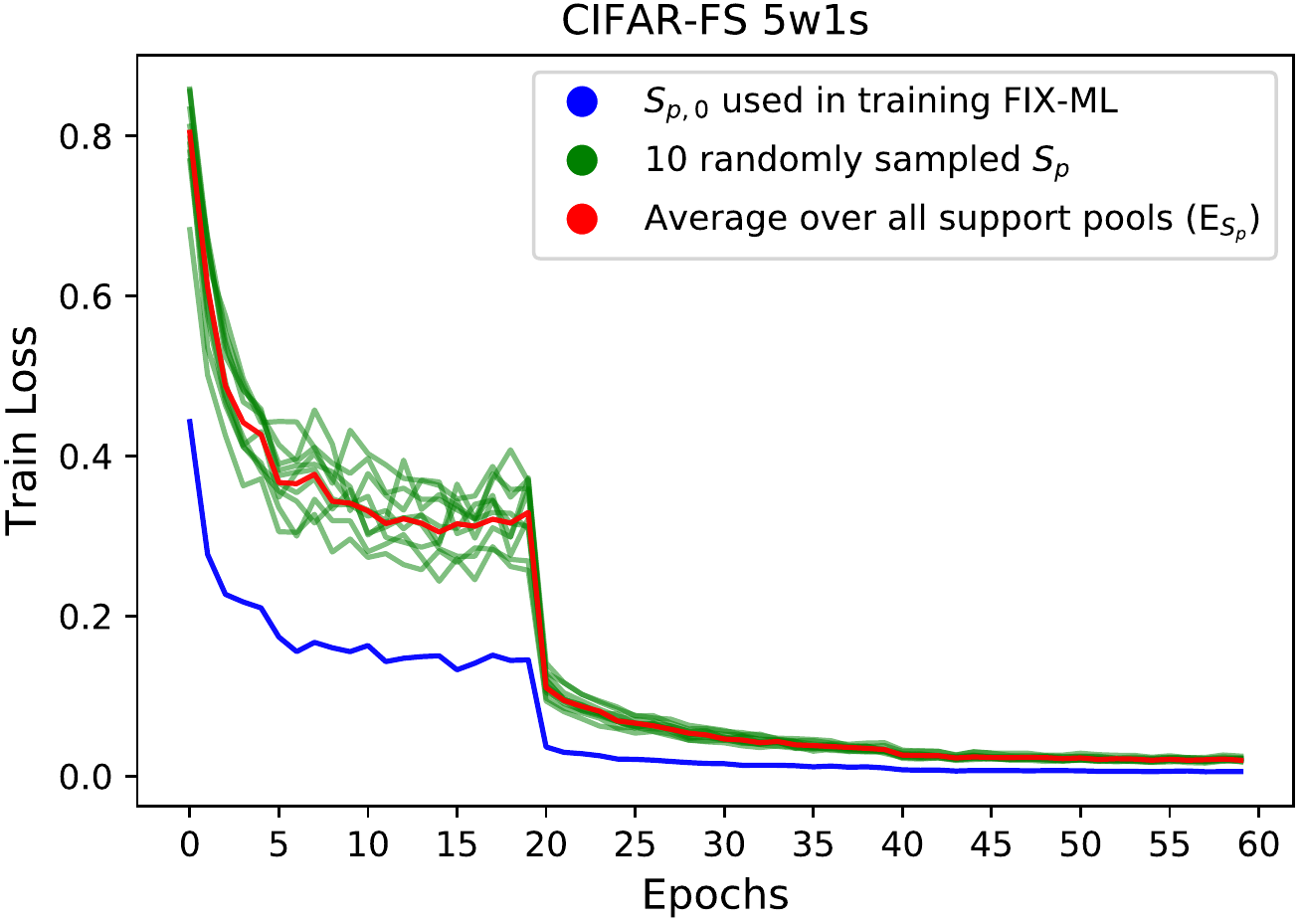} 
    \caption*{(c)}
  \end{minipage} 
  \vspace{1em}
 \caption{In (a) we compare \fixml learned Protonets with other popular methods\protect\footnotemark\xspace on \mini-5w5s. On \cif-5w5s (b) and \cif-5w1s (c) we evaluate the \fixml-learned algorithm after each training epoch using three criterions; \textit{blue}: \fixml's loss on the specific support pool ($S_{p,0}$) used in its objective \eqref{eq:fixML}; \textit{red}: the meta-learning objective in \eqref{eq:ML}[B] and \textit{green}: the objective in \eqref{eq:fixML} but on 10 randomly sampled support pools different from $S_{p,0}$.\vspace{1em}}
   \label{fig:2} 
\end{figure}
\footnotetext{For MetaOptNet-SVM we report the performance observed by \cite{lee2019meta} without label smoothing.}

Naively, one might think that because \fixml is optimized on a \textbf{single} fixed support pool, the training loss on any other support pool would be much worse.   To verify this idea, 
we re-evaluate the \fixml-learned algorithms \textbf{1)} over $10$ other fixed support pools by replacing $S_{p,0}$ in \eqref{eq:fixML} with $S_p \neq S_{p, 0}$; \textbf{2)} on the original meta-learning objective \eqref{eq:ML}[B] by averaging over all possible support pools $S_p$ (because of the large number of possible support pools, we estimate this average using a very large number of sampled tasks). We conduct this evaluation for Protonet on \cif-5w5s and \cif-5w1s tasks, for all the trained algorithm snapshots over the entire $60$ training epochs. Here, the random class set $C$ in \eqref{eq:ML}[B] is set to always contain exactly $5$ classes ($|C| = 5$) in order to match the meta-test scenario. The results are shown in Figure~\ref{fig:2}(b),(c). As we see in both plots, despite optimizing for one fixed support pool (blue curve), the training losses of other support pools (green curves) and the \ml training loss (red curve) are all decreasing consistently in the same trend as the blue curve throughout the entire optimization trajectory. 
To understand this consistent decrease, we note that the gradient of \fixml objective is a biased gradient of the original \ml objective. When the gradient bias is not too large, stochastic gradient descent using a biased gradient can still lead to a reduction in objective loss and convergence 
\cite{bottou2018optimization}, which is what we observe for the \ml objective when using biased stochastic \fixml gradients.

\begin{figure}
    \centering
    \includegraphics[width=\linewidth]{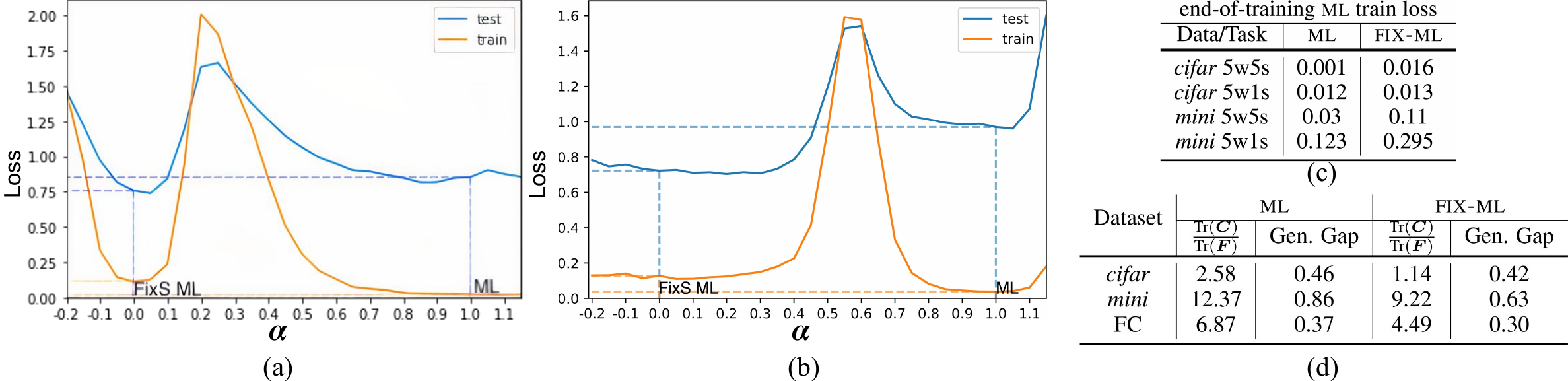}
    \caption{In (a),(b) (two different runs) we plot the train and test loss (averaging over all support sets) for network weights obtained by linearly interpolating between the \fixml ($w_{fml}$ at $0.$) and \ml ($w_{ml}$ at $1.$) Protonet solutions for \mini-5w5s tasks. We evaluate \eqref{eq:ML}[B] with $w = (1-\alpha)w_{fml} + \alpha w_{ml},\,\, \alpha \in [-0.2, 1.2]$. In (c) we compare the \ml train losses for \fixml and \ml learned Protonets at the end of training. In (d) we show the correlation between the generalization gap and {\scriptsize $\frac{\tr(\bC)}{\tr(\bF)}$} on 5w5s tasks for \cif, \mini (Protonet) and \FC\xspace(SVM).}
    \label{fig:3}
\end{figure}

However, this does not give us a complete picture for meta-training. Even though \ml training loss is considerably reduced for the \fixml-learned algorithm, we need to compare this value against \ml-training loss of the corresponding \ml-learned algorithm to understand the meta-test performance difference. In Figure \ref{fig:3}(c), we compare the end-of-training \fixml solutions with the corresponding \ml solutions using the \ml training objective \eqref{eq:ML}[B] on 4 different dataset/task configurations. We see that \fixml still consistently achieves higher \ml-training losses than their \ml counterparts. With a relatively lower \ml test loss but a much higher \ml training loss, the \fixml solution must have a smaller generalization gap than that of the \ml solution.
Since there isn't a single best way to understand neural networks' generalization gap, we look at some of the recently proposed metrics:


\textbf{1) Sharpness of the training loss landscape:} A  popular hypothesis (dating back to \cite{hochreiter1997flat}) is that the flatness of the training loss landscape is correlated with (and possibly leads to) a smaller generalization gap \cite{keskar2016large, chaudhari2019entropy}. Several sharpness metrics based on this hypothesis were called into question by \cite{dinh2017sharp}, showing that a functionally equivalent neural network with differently scaled parameters could have arbitrarily different sharpness according to these metrics. However, it was argued in \cite{jastrzkebski2017three} that even though sharp minima with similar test performance as flat minima do exist, stochastic gradient descent will not converge to them. While further understanding sharpness's relationship with generalization gap is still an ongoing research topic, we still consider two sharpness analyses here.

\textbf{1a) Sharpness visualization using 1d interpolation plots} was first proposed by \cite{keskar2016large} and also used by \cite{izmailov2018averaging} to compare the flatness of two different solutions' training loss landscape when explaining the two solutions' generalization gap differences. For two models with the same neural architecture but different weights, training and test loss are evaluated at model parameters that are different linear interpolations of the two models' weights. In our case, we interpolate between the parameters of end-of-training \fixml-learned algorithm and \ml-learned algorithm from Protonet for miniImagenet 5w5s. This result is shown in Figure \ref{fig:3}(a),(b). In \ref{fig:3}(a), we see that \fixml is located at a relatively ``sharper'' minima than \ml despite having a smaller generalization gap. This is inconsistent with observations in \cite{keskar2016large, izmailov2018averaging} possibly because in our case, the loss landscape between meta-train and meta-test are \textbf{not} necessarily shifted versions of each other (confirmed also in Figure \ref{fig:3}(b)). Instead, we see that the trends of these two graphs align well with each other. Therefore, we cannot solely rely on 1d visualization-based sharpness analysis to explain the generalization gap.


\textbf{1b)} In addition to visualizations, \textbf{quantitative measures of the training loss landscape sharpness} has also been used. Specifically, one notion of the loss landscape sharpness is in the curvature of the loss function. This is typically measured by the maximum eigenvalue \cite{li2018visualizing} or the trace  $(\textrm{tr}(\mathbf{H}))$ of the Hessian matrix ($\mathbf{H}$) \cite{jastrzkebski2017three} for the parameters at the end of training . However, directly computing $\mathbf{H}$ accurately is both computationally expensive and memory-wise ``impossible" for over-parameterized models like Resnet-12. Thus, $\mathbf{H}$ is typically approximated with the Fisher information matrix ($\mathbf{F}$) at the same parameter \cite{thomas2020interplay}.  $\textrm{tr}(\mathbf{F})$ can be easily estimated without constructing the entire matrix $\mathbf{F}$. However, $\mathbf{F}$ is only equal to $\mathbf{H}$ when the parameters are the true parameters of the training set (in practice, achieve training loss extremely close to $0$). On the other hand, this approximation is very crude when the training loss is much larger than $0$ in the case of \fixml (Figure~\ref{fig:3}(c)). Hence, $\textrm{tr}(\bH)$ for \fixml's parameters cannot be accurately estimated using $\tr(F)$. As a result, for our analysis, this approach is not readily applicable. 

\textbf{2) Another quantitative measure which approximates the Takeuchi Information Criterion (TIC)} \cite{takeuchi1976distribution} was empirically shown to correlate more with the generalization gap as compared to measures of the loss landscape's flatness \cite{thomas2020interplay}. Specifically, \citeauthor{thomas2020interplay} (\citeyear{thomas2020interplay}) proposes the quantity {\scriptsize $\frac{\tr(\bC)}{\tr(\bF)}$}, where $\bC$ is the uncentered covariance matrix of the gradient with respect to the training data distribution. To evaluate how well this measure tracks the generalization gap difference between \fixml and \ml, we evaluate this approximate TIC value for three pairs of \ml and \fixml-learned algorithms 
in Figure~\ref{fig:3}(d). We see that for all three datasets, \ml has both higher {\scriptsize $\frac{\tr(\bC)}{\tr(\bF)}$} and larger generalization gap than its \fixml counterpart, indicating some possible correlation between these two. Thus it would be interesting to understand what intuitive property of the solutions is captured by this quantity and why optimizing the \fixml objective leads to solutions with lower {\scriptsize $\frac{\tr(\bC)}{\tr(\bF)}$}.

\vspace{-0.5em}



\section{Conclusion}
\vspace{-1em}
In this paper, we have proposed a simple yet effective modification of the meta-learning objective. We begin by re-formulating the traditional meta-learning objective, which motivates our approach of reducing the support set diversity. 
Then, we experimentally show that for a variety of meta-learning methods, datasets, and model architectures, the algorithms learned using our objective result in superior generalization performance relative to the original objective.  
We analyze this improvement by understanding the optimization loss and the generalization gap separately. 
While the optimization loss trade-off is easier to reason about using our framework, the generalization gap is still not fully understood based on several initial analyses, which we hope would pave the way for future work in this area.

\bibliographystyle{abbrvnat}
\bibliography{references}

\clearpage
\end{document}